\documentclass{article}
\usepackage{spconf,amsmath,graphicx, multirow, dsfont, cite}

\usepackage{psfrag}
\usepackage{fancyhdr}
\usepackage[absolute]{textpos}

\title{Semi-Supervised 3D Object Detection with Channel Augmentation using Transformation Equivariance}

\name{Minju Kang$^{\star \ddagger}$ \qquad Taehun Kong$^{\star}$ \qquad Tae-Kyun Kim$^{\star \dagger}$}
\address{$^{\star}$KAIST, $^{\dagger}$Imperial College London, $^{\ddagger}$LG Electronics}

\begin{document}

\begin{textblock*}{20cm}(0.5cm,0.5cm)
    \noindent
    \centering
    \footnotesize
    \copyright 2024 IEEE. Personal use of this material is permitted. Permission from IEEE must be obtained for all other uses, in any current or future media, including reprinting/republishing this material for advertising or promotional purposes, creating new collective works, for resale or redistribution to servers or lists, or reuse of any copyrighted component of this work in other works
\end{textblock*}

\maketitle

\let\thefootnote\relax\footnotetext{This work was in part sponsored by NST grant (CRC 21011, MSIT), IITP grant (RS-2023-00228996, MSIT) and KOCCA grant (R2022020028, MCST).
}
\begin{abstract}
Accurate 3D object detection is crucial for autonomous vehicles and robots to navigate and interact with the environment safely and effectively. Meanwhile, the performance of 3D detector relies on the data size and annotation which is expensive. Consequently, the demand of training with limited labeled data is growing.
We explore a novel teacher-student framework employing channel augmentation for 3D semi-supervised object detection. The teacher-student SSL typically adopts a weak augmentation and strong augmentation to teacher and student, respectively. In this work, we apply multiple channel augmentations to both networks using the transformation equivariance detector (TED). The TED allows us to explore different combinations of augmentation on point clouds and efficiently aggregates multi-channel transformation equivariance features. In principle, by adopting fixed channel augmentations for the teacher network, the student can train stably on reliable pseudo-labels. Adopting strong channel augmentations can enrich the diversity of data, fostering robustness to transformations and enhancing generalization performance of the student network. We use SOTA hierarchical supervision as a baseline and adapt its dual-threshold to TED, which is called channel IoU consistency. We evaluate our method with KITTI dataset, and achieved a significant performance leap, surpassing SOTA 3D semi-supervised object detection models.
\end{abstract}

\section{Introduction}
\label{sec:intro}

With the growing demand of relevant applications e.g. autonomous vehicles, research on 3D object detection becomes increasingly important. While recent progress in 3D object detection is impressive, further improving performance demands a large-scale dataset and accurate instance-level annotations. Generating such 3D labels is considerably costly, which emphasizes the critical need of robust semi-supervised learning techniques to alleviate the resource.

Semi-supervised learning (SSL) encompasses two primary paradigms: consistency regularization and pseudo-labeling. Consistency regularization \cite{temporal, ts, ladder, vat} aims to improve model generalization by encouraging consistent predictions for the same input data under different perturbations. Pseudo-labeling \cite{pseudo, noisy, prop, simclrv2} selects the model-generated prediction which has the maximum probability and exploits them as labels. Recent research \cite{mixmatch, remixmatch, fixmatch} has achieved significant performance gains in semi-supervised learning by effectively combining these two strategies. This often employs a teacher-student framework where the two models utilize different data augmentation intensities. The teacher model generates pseudo-labels for unlabeled data with a weak augmentation (e.g., flip, translation, crop). Subsequently, the student model is trained on both labeled and pseudo-labeled data, typically employing a stronger augmentation (e.g., Cutout \cite{cutout}, RandAugment \cite{randaugment}, CTAugment \cite{remixmatch}).

The key point of semi-supervised object detection (SSOD) is transformation robustness. Compared to 2D images, 3D point clouds have inherent challenges for interpreting and understanding the scene due to increased dimensions and varying point density. These complexities hinder the reliable predictions of a teacher model with weakly-augmented data. To address these challenges, we adopt channel augmentation and transformation equivariant detector (TED) \cite{ted} for teacher and student network. For clarity, weak and strong augmentation in previous SSL works do not mean increasing the amount of data itself but modify the data through transformation. On the other hand, the channel augmentation generates multiple transformed point clouds as input from the original point clouds. The multi-channel point clouds are processed at once by TED thus being efficient than naively augmenting the size of training data. TED extracts voxel features for each distinct channel and aggregates and aligns them so that the model can learn transformation equivariant features. By considering multiple transformed inputs, the teacher model is less likely to fixate on specific features or patterns that might be present in a single, untransformed view. Furthermore, the strong channel augmentation for student model effectively expands the dataset with diverse transformations. With the broader scenes, student TED fosters robustness to transformations which is important in consistency regularization based SSL.

To evaluate the efficacy of channel augmentation in the context of 3D SSOD, we employ HSSDA \cite{hssda} as a SOTA baseline. HSSDA stratifies pseudo-boxes based on their classification confidence, objectness and IoU consistency. The detector directly outputs the classification and objectness score, whereas IoU consistency requires a distinct calculation to measure box localization quality under consistency constraint. The boxes generated from original scenes are matched with maximal overlap within the predicted boxes from weakly-augmented scenes. Since TED outputs channel-wise box predictions from each voxel feature while sharing RoI predicted from the aggregated features, we eliminate the need for additional forward processing or matching steps. Instead, we leverage the average of IoU across channel-wise box predictions to effectively evaluate box quality.

Our contributions can be summarized as follows:
\begin{itemize}
    \item To emphasize the importance of diverse data and transformation equivariance in SSL, we inject  channel augmentations to teacher-student framework.
    \item To supervise our network with reliable pseudo-boxes, we average channel-wise predictions and use their IoU for filtering criteria.
    \item Our method significantly outperforms existing SOTA methods on KITTI validation dataset and we evaluate the incremental performance gains of channel augmentation and filtering method.
\end{itemize}

\section{Related Work}
\label{sec:related}

\subsection{Semi-Supervised Learning}
\label{ssec:ssl}
Among various semi-supervised techniques, consistency regularization \cite{temporal, ts, ladder, vat} and pseudo-labeling \cite{pseudo, noisy, prop, simclrv2} have emerged as particularly successful methodologies. For consistency regularization method, UDA \cite{uda} shows that learning consistency between the outputs of applying weak and strong data augmentation can outperform prior methods. Fixmatch \cite{fixmatch}, by combining UDA and pseudo-labeling, demonstrates impressive performance across a wide range of datasets. They demonstrate the importance of weak augmentation by conducting the experiment replacing weak augmentation with no augmentation. As the result, the model overfits the guessed unlabeled labels and gets lower performance. In object detection task, \cite{stac} also adopts this weak-strong augmentation scheme and filters pseudo-boxes with the confidence score. Several works focus on improving localization quality of pseudo-boxes. These include \cite{bmvc} guided by aleatoric uncertainty and Softeacher \cite{softteacher} that leverages the variance of iteratively refined boxes. While most of these methods find optimal transformation type and its magnitude for strong augmentation with \cite{randaugment, remixmatch}, our strong channel augmentation enables us to explore various combinations of transformation magnitude.

\subsection{3D Semi-Supervised Object Detection}
\label{ssec:3dssod}
Recent works for 3DSSOD have explored domain-specific techniques. SESS \cite{sess} designs three consistency losses to align object locations, semantic categories and sizes predicted by the teacher and student network. Most recent 3DSSOD focus on generating reliable pseudo-boxes. 3DIoUMatch \cite{3dioumatch} employs 3D IoU as the primary criterion for mining pseudo-boxes, contributing to enhanced localization quality. DetMatch \cite{detmatch} matches 2D and 3D detections to generate cleaner pseudo-boxes, compensating for modality-specific weaknesses. Proficient Teacher \cite{proficient} integrates predicted boxes from fixed augmented multiple point clouds to enhance recall, and ensures higher precision by its learnable box voting module. Both our method and the Proficient Teacher utilize multiple point clouds using fixed weak augmentation. However, Proficient Teacher needs multiple forward processing and additional post-processing to deviate from conventional approach that rely heavily on threshold selection while we enhance efficiency by using TED \cite{ted} and focus on refining the threshold. DDS3D \cite{dds3d} proposes dense pseudo-label generation rather than NMS which can remove beneficial boxes. HSSDA \cite{hssda} employs hierarchical supervision based on dual-threshold, yielding a substantial improvement in detection performance. In addition, novel shuffle data augmentation strengthens the existing strong augmentation for 3D point clouds. In contrast to NoiseDet's \cite{noisedet} that focuses on BEV feature consistency with two strongly augmented scenes, our method uses TED \cite{ted} to enforce consistency of channel-wise outputs for extracting transformation equivariant features on every module.

\section{SSL using TED and hierarchical supervision}
\label{sec:method}

\begin{figure*}[t!] 
\centering
  \includegraphics[width=\textwidth]{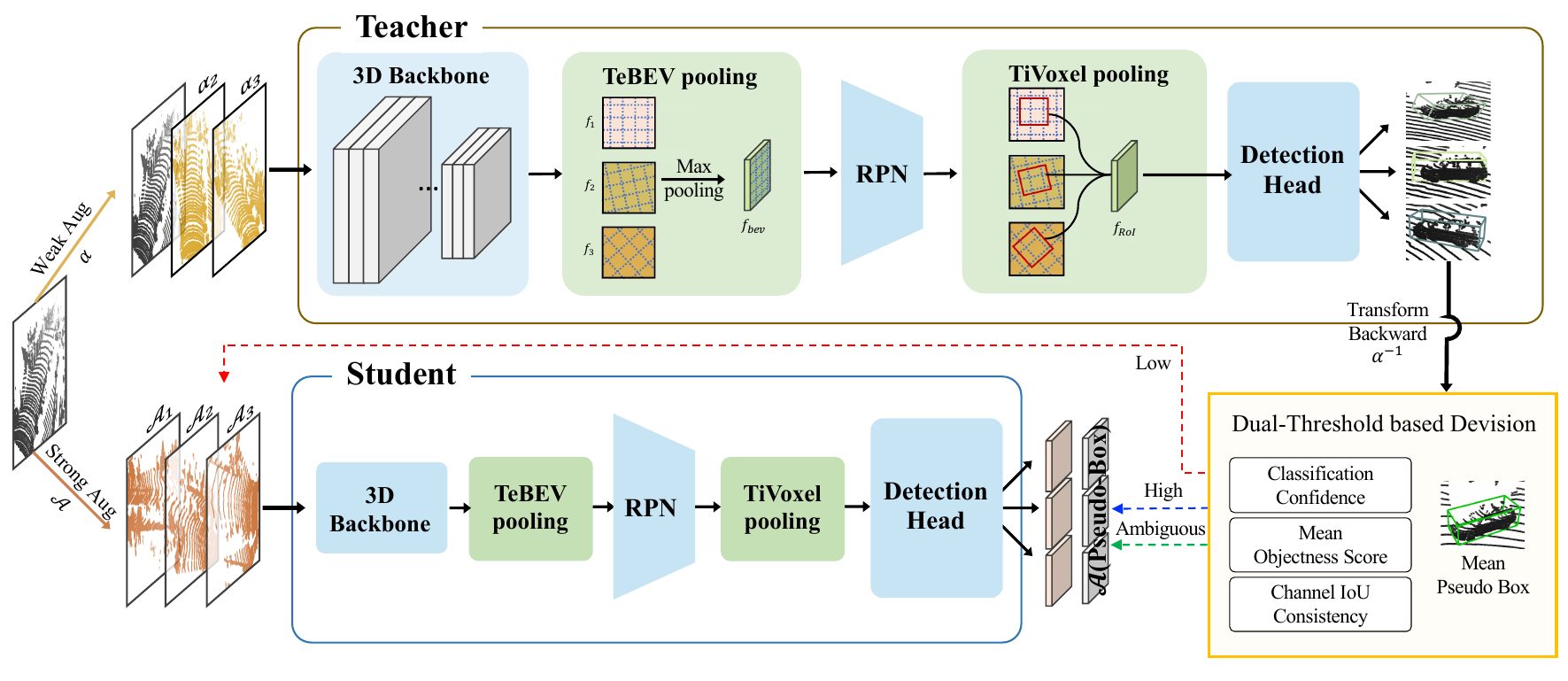}
\caption{Overview of the proposed method. It augments input channels to the teacher and student and aggregates them using transformation equivariance features as in TED. HSSDA is applied with the pseudo-box qualities based on TED.}
\label{fig:main}
\end{figure*}

\subsection{Method Overview}
Figure \ref{fig:main} illustrates our overall framework. Unlike other methods, we input augmented multiple point clouds together and process it by TED \cite{ted}. We control the intensity of the data augmentation with randomness: fixed for the teacher, random for the student. The teacher outputs multiple box residuals for one object using RoI feature of each channel. By averaging the box coordinates, we use it as a pseudo-box to supervise the student. To assess its localization quality, we calculate the IoU with the pairs of box predictions. Employing this channel IoU consistency, we categorize pseudo-boxes into distinct levels. Then pseudo-boxes excluding low level are transformed with the parameters of strong channel augmentation to explicitly model the transformation robustness of student TED. The preliminary information of TED is described in Section \ref{ssec:channel} and our training method is detailed in Section \ref{ssec:train} and Section \ref{ssec:obj}.
              
\subsection{Notations for the teacher-student SSL}
\label{ssec:def}
For semi-supervised 3D object detection, labeled data $\boldsymbol{D}^s=\{\boldsymbol{x}^s_i, (\boldsymbol{b}_i^*, \boldsymbol{c}_i^*)\}^{N^s}_i$ and unlabeled data $\boldsymbol{D}^u=\{\boldsymbol{x}^u_i\}^{N^u}_i$ are used in training. The ground-truth boxes in each labeled data are annotated as $\boldsymbol{b}_i^*=\{(cx_{ij},cy_{ij},cz_{ij},w_{ij},h_{ij},l_{ij},r_{ij}) \in \mathcal{R}^7 \}^{N^{B_i}}_j$, which represent box center coordinates, size and orientation in corresponding order. Another annotation $\boldsymbol{c}^*_i=\{c_{ij}\}^{N^{B_i}}_j$ is a set of class indices for every boxes. For teacher and student network, weak augmentation $\alpha$ and strong augmentation $\mathcal{A}$ is applied to teacher model and student model, respectively. The teacher model's weight is updated via exponential moving average (EMA) of the student model's weights following \cite{meanteacher}.

\subsection{Background: Transformation Equivariant Detector}
\label{ssec:channel} 
We utilize TED \cite{ted} as a detector in teacher-student framework to enhance the robustness to transformation which is critical in SSL. To explicitly model transformation equivariance, TED fixes transformation actions $\{\mathcal{T}_i\}^{N_C}_i$ and transforms point clouds into $N_C$ distinct point clouds. TED adopts Voxel-RCNN \cite{voxelrcnn} as its structural baseline, which is composed of 3D backbone network, 2D region proposal network (RPN) and detection head.

At first, each of the point clouds is encoded to multi-level voxel features $\mathcal{F}_i$ by the backbone network.
Then, the voxel features are converted into BEV features $\mathcal{F}^{BEV}_i$ by compressing along height dimension. To align across the transformation channels, grid points $G$ are generated under $\mathcal{F}_1^{BEV}$ space, serving as the basis for feature interpolation. The aligned features are subsequently max-pooled, resulting in the efficient generation of BEV feature representation $\mathcal{F}^{BEV}$.
\begin{equation}
 \mathcal{F}^{BEV} = \mathcal{M}(\{\mathcal{I}(\mathcal{F}_i^{BEV}, \mathcal{T}_i(\mathcal{T}_1^{-1}(G)))\}^{N^C}_i)
\end{equation}
RPN takes the unified BEV feature as input and creates both box proposals $B$ and classification confidences.
After the NMS of box proposals, each of the RoI features is extracted via pooling operations from corresponding voxel features.
\begin{equation}
 \mathcal{F}^{RoI}_i =\text{Pool}(\mathcal{F}_i, \mathcal{T}_i(\mathcal{T}_1^{-1}(B))), i=1,...,N^C
\end{equation}
Detection head generates bounding box predictions and objectness scores from RoI features.
The final box predictions and objectness scores are derived by averaging all predictions which are transformed backward and scores, respectively.

\subsection{Learning with Transformation Channels}
\label{ssec:train}
By using TED \cite{ted} as a detector during semi-supervised learning, we emphasize the advantages of using transformation channel in SSL. We adapt the transformation intensity based on the model, employing fixed transformation for the teacher model, $M_T(\{\alpha_i(x)\}^{N^C}_i)$, and random augmentation for the student model, $M_S(\{\mathcal{A}_i(x)\}^{N^C}_i)$. As a SSL baseline, we deliberately leverage 3-step hierarchical supervision proposed by HSSDA \cite{hssda}, which is the state-of-the-art method.

\begin{figure}[htb]

\begin{minipage}[b]{1.0\linewidth}
  \centering
  \centerline{\includegraphics[width=8cm]{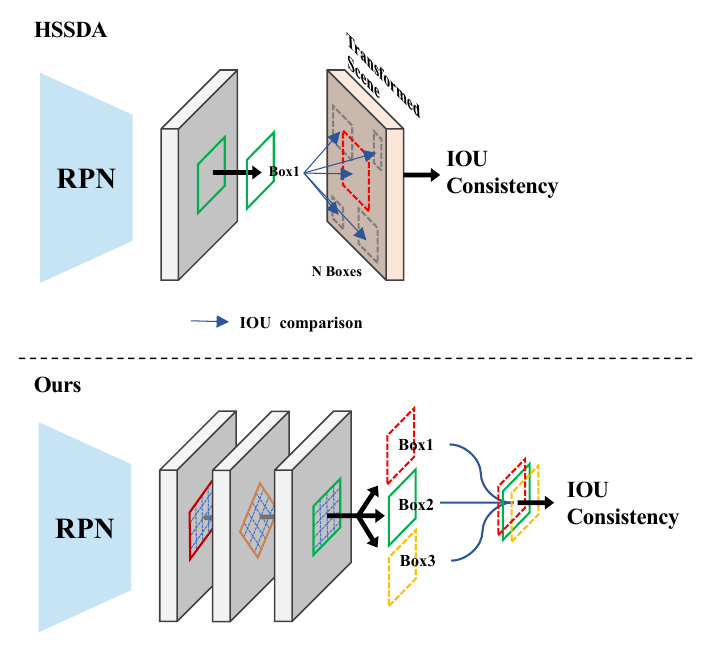}}
\end{minipage}
\caption{IoU consistency comparison.}
\label{fig:iou}
\end{figure}

Following HSSDA, the first step is to generate dual-threshold with confident scenes.  We apply $N_C$ channel augmentations with fixed parameters which is called weak-augmentation $\alpha$ to the scenes. Then the output of the teacher is a set of $N_C$ bounding boxes, classification confidences, and averaged objectness scores. To measure the quality of the pseudo-boxes, HSSDA uses the classification confidence score, objectness score and IoU consistency of each predicted box as a criteria. By these components, boxes are clustered into three groups and the two boundaries are used as dual-threshold. We retain the classification confidence and objectness score to adhere to the established framework. Additionally, we calculate IoU along the predicted boxes of transformation channels $\alpha_i^{-1}(B_i)$ which is illustrated in Figure \ref{fig:iou}. The following advantages distinguish our channel IoU consistency method: (1) Our approach eliminates the need for any additional forward processing. The original approach necessitates two separate forward processes for IoU consistency calculation: one for the original point cloud and another for its weakly augmented counterpart. By using TED \cite{ted}, $N_C$ data inferences converge into a unified process, streamlining the overall pipeline. (2) We achieve computational savings on calculating IoU. HSSDA calculates IoU between $N_1$ predicted boxes predicted from the normal point clouds and $N_2$ boxes from weakly-augmented ones, which requires $O(N_1 \cdot N_2)$ complexity. Then, the two sets of boxes are paired based on maximum overlapping criteria to estimate IoU consistency of the boxes. In contrast, the shared RoI among the $N_C$ boxes where $N_C$ is a small constant enables us to eliminating the pairing process of HSSDA, leading to linear $O(N_1)$ complexity.

In the second step, pseudo-boxes are generated by the teacher network and subsequently stratified into three levels: high, ambiguous and low level, using dual-thresholds. Notably, our method does not require two distinct inference processes as previous step. In the last step, student network is trained with hierarchical supervision. To avoid the detection at the background, points inside the low level pseudo-boxes are removed from the input of the student model. We apply random channel augmentation $\mathcal{A}$ and the augmented samples are used to predict $N_C$ boxes for each RoI. As same as the original, the predicted boxes which are assigned to high-confidence level pseudo-boxes are treated as labeled data, while the ambiguous level boxes are supervised in soft-weight manner. We also adopt shuffle data augmentation but omitted in Figure \ref{fig:main} for visual clarity.

\subsection{Training Objectives}
\label{ssec:obj} 
We pretrain Voxel-RCNN which has same learnable parameters with TED to align the pretraining method with existing methods. The teacher network is initialized by pretrained Voxel-RCNN and updated by EMA. Note different pretraining lead to different accuracies (see implementation details in Section 4.1). The total loss for the student network is calculated by the sum of RPN and detection head losses.
\begin{equation}
 \mathcal{L}_{total}=\mathcal{L}_{rpn}^s + \mathcal{L}_{rpn}^u + \mathcal{L}_{head}^s + \mathcal{L}_{head}^u
\end{equation}
where $\mathcal{L}^s_{rpn}$ and $\mathcal{L}^s_{head}$ follow original TED \cite{ted} losses. The losses for unlabeled data are explained in details. 

{\bf RPN.}
For every anchor, classification loss is computed while regression loss is not at background anchor (i.e., $c^*_i=0$). 
\begin{equation}
 \mathcal{L}_{rpn}^u = \sum_i^{N^{a}} w_i \mathcal{L}_{cls}(p_i,c_i^*) + w_i \mathds{1}(c_i^*>1) \mathcal{L}_{reg}(\Delta b_i^a, \Delta \hat{b}_i^a)
\end{equation}
where $\Delta b_i^*$ is encoded by the residual of $b_i^*$ and $b_i^a$.

{\bf Detection head.}
$N_{roi}$ is the number of RoIs where the predicted class label matches the ground-truth label.
\begin{equation}
 \mathcal{L}_{head}^u = \sum_i^{N^{roi}} \sum_j^{C} w_i \mathcal{L}_{cls}(o_{ij}, \hat{c}_i) + w_i \mathcal{L}_{reg}(\Delta b_{ij}^r, \Delta \hat{b}_i^r)
\end{equation}
\begin{equation}
  w_i =
    \begin{cases}
      0 & \text{if $b_i^r$ in low-level} \\
      \hat{p}_i \times \hat{o}_i & \text{if $b_i^r$ in ambiguous-level} \\
      1 & \text{if $b_i^r$ in high-level}
    \end{cases}
\end{equation}
$\hat{c}$ and $\hat{b}$ are class label and coordinate of pseudo-box, respectively.

\section{Experiments}
\label{sec:}

\begin{table*}[htb!] 
\centering
\begin{center}
\begin{tabular}{ c|c|cccc|cccc|cccc } 
 \hline
  \multirow{2}{*}{Model} & \multirow{2}{*}{Modality} & \multicolumn{4}{c|}{1\%} & \multicolumn{4}{c}{2\%} & \multicolumn{4}{|c}{20\%}\\
  & & Car & Ped & Cyc & Avg & Car & Ped & Cyc & Avg & Car & Ped & Cyc & Avg\\
 \hline
 \hline
 *PV-RCNN \cite{pvrcnn} & LiDAR & 73.5 & 28.7 & 28.4 & 43.5 & 76.6 & 40.8 & 45.5 & 54.3 & 77.9 & 47.1 & 58.9 & 61.3 \\ 
 \hline 
 3DIoUMatch \cite{3dioumatch} & LiDAR & 76.0 & 31.7 & 36.4 & 48.0 & 78.7 & 48.2 & 56.2 & 61.0 & - & - & - \\ 
 \hline
 DetMatch \cite{detmatch} & LiDAR+RGB & 77.5 & \bf{57.3} & 42.3 & 59.0 & 78.2 & 54.1 & 64.7 & 65.6 & 78.7 & 57.6 & 69.6 & 68.7 \\
 \hline
 HSSDA \cite{hssda} & LiDAR & 80.9 & 51.9 & 45.7 & 59.5 & 81.9 & 58.2 & 65.8 & 68.6 & 82.5 & {\bf 59.1} & {\bf 73.2} & 71.6 \\
 \hline
 \hline
 *Voxel-RCNN \cite{voxelrcnn} & LiDAR & 72.6 & 26.2 & 30.4 & 43.1 & 76.0 & 39.0 & 44.8 & 53.3 & 79.6 & 43.6 & 61.8 & 61.7 \\
 \hline
 *TED \cite{ted} & LiDAR & 72.4 & 23.3 & 33.1 & 42.9 & 75.6 & 39.3 & 41.5 & 52.1 & 77.4 & 39.4 & 55.6 & 57.5 \\
 \hline

 Ours & LiDAR & \underline{\bf 82.4} & \underline{57.0} & \underline{\bf 56.7} & \underline{\bf 65.4} & \underline{\bf 82.8} & \underline{\bf 61.0} & \underline{\bf 72.1} & \underline{\bf 72.0} & \underline{\bf 83.7} & \underline{57.0} & \underline{72.1} & \underline{70.9} \\
 \hline
\end{tabular} 
\end{center}
\caption{3D semi-supervised object detection performance comparison on KITTI dataset. An asterisk* indicates pretrain models. The reported number of TED which we use as the pretrained model is the test result when TED is initialized with pretrained Voxel-RCNN. Among all models, the highest performances are in the bold font. The underlined results improved most by each of the semi-supervised methods.}
\label{tab:kitti}
\end{table*}

\subsection{Dataset and Metrics}
\label{ssec:dataset}
{\bf KITTI Dataset.} The KITTI 3D object detection benchmark \cite{kitti} consists of 3,712 training frames and 3,769 validation frames which are used for evaluation. We follow the labeled data generation and evaluation metrics of the previous works \cite{hssda, 3dioumatch}. For semi-supervised training, each of 1\%, 2\% and 20\% labeled set is sampled from training frames and the remaining frames are used as the unlabeled set. We calculate the mAP with 40 recall positions and report the average of 3 labeled data splits. The prediction boxes of the model are considered as true positive when the 3D IoU with the ground-truth boxes are over 0.7, 0.5, and 0.5 for the three classes: car, pedestrian, and cyclist, respectively.

\subsection{Implementation Details}
\label{ssec:details}
{\bf Data Processing.} We define the channel number $N_C$ as 3. For weak channel augmentation, we use original point cloud without any perturbation for the first channel and fix the transformation scale for the others. The other two channels use flipped scene and transform it by rotation with -22.5$^\circ$ and 22.5$^\circ$ degree, and scale factor of 0.98 and 1.02 for each. For strong channel augmentation, we randomly flip the scene, rotate within a range of -45$^\circ$ to 45$^\circ$, scale within a range of 0.95 to 1.05.

{\bf Network Architecture.} We use Voxel-RCNN \cite{voxelrcnn} for pretraining the model. For semi-supervised training, we use TED \cite{ted}, which is based on Voxel-RCNN. They use an attention layer in the detection head to aggregate multiple features for the final prediction. We omitted the attention layer to align the number of learnable parameters within our framework with that of Voxel-RCNN, enabling initializing teacher and student network with the pretrained network. This pretraining gave us a considerable performance gain. 

{\bf Training Details.} All pretrained Voxel-RCNNs are obtained after 80 epochs with a batch size 16. Subsequently, we trained TED for 80 epochs with a batch size of 8 using two 3090 GPUs. We used ADAM as an optimizer. The dual threshold is generated every 5 epochs following HSSDA \cite{hssda}.

\subsection{Main Results}
\label{ssec:result}

\begin{figure*}[t!]
\centering
\includegraphics[width=\textwidth]{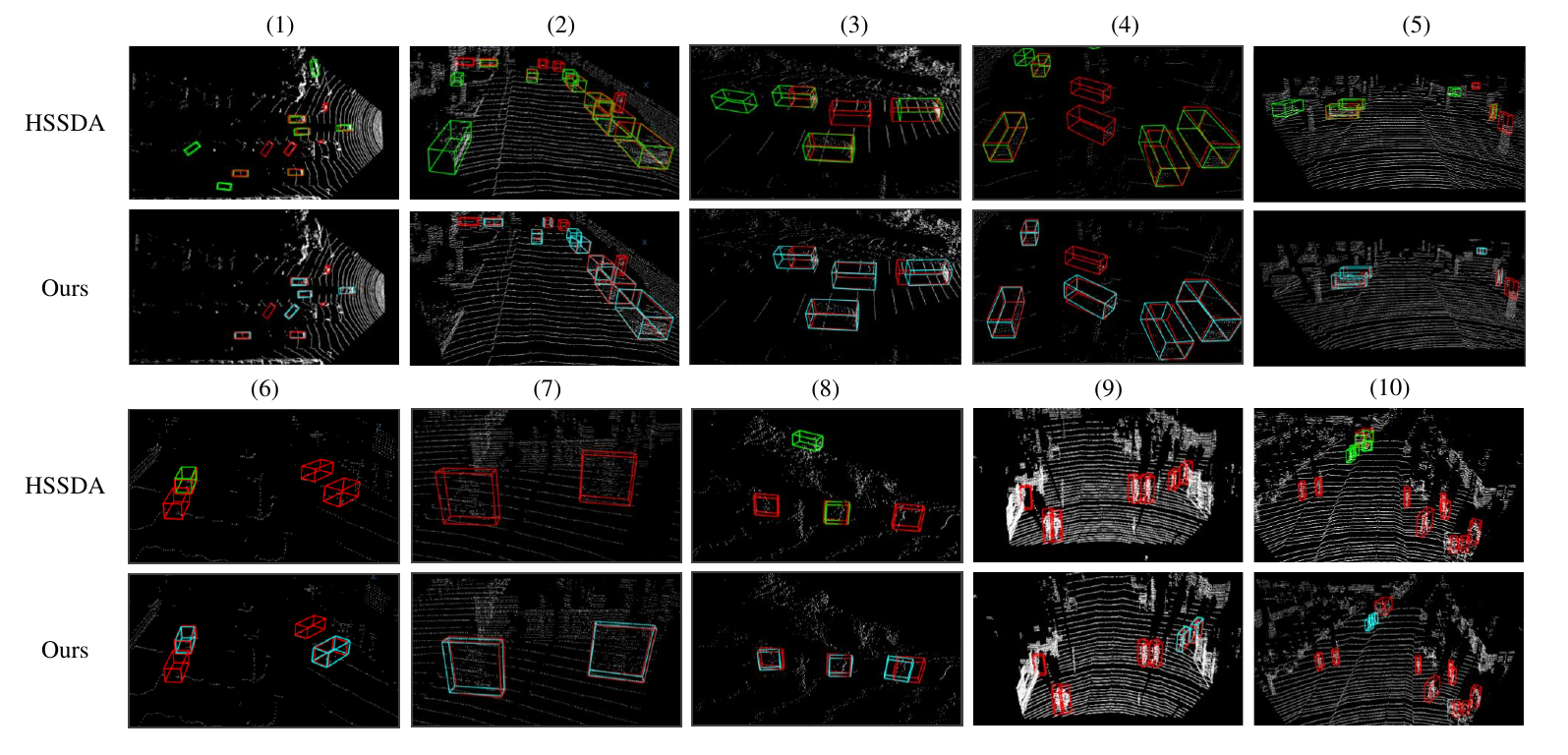}
\caption{Qualitative comparisons of pseudo-boxes on KITTI. Ground truth bounding boxes appear in red, our predicted pseudo-boxes in cyan, and HSSDA's pseudo-boxes in green.}
\label{fig:results}
\end{figure*}

We compare our method with other state-of-the-art methods on KITTI val set. As shown in Table \ref{tab:kitti}, our framework demonstrated remarkable performance gains, particularly in scenarios with extremely limited labeled data. Using 1\% and 2\% labeled data, our model significantly surpassed the pretrained TED model by a margin of 22.5\% and 19.9 \% mAP, respectively. We outperform DetMatch's 15.5\% and 11.3\% improvement and HSSDA's 16.0\% and 14.3\% gain over their PV-RCNN pretrained models. This performance gain was achieved without adopting any additional modalities e.g. RGB or learnable parameters, highlighting the efficiency and robustness of our approach. In 20\% labeled data, since our TED model is initialized with Voxel-RCNN and tested with fixed channel augmentation, the performance is lower than the others. However, our method beats other works on improvement over pretrained models.
We also compare the quality of pseudo-boxes with HSSDA. As shown in Figure \ref{fig:results}. (1)-(8), our method minimizes false positives, maintains robust performance on transformed objects, and excels in the challenging task of cyclist detection. However challenges of detecting small objects like pedestrian remains (see Figure \ref{fig:results}. (9), (10)).

\subsection{Ablation Study}
\label{ssec:ablation}
\begin{figure}[htb]
 \centering
 \centerline{\includegraphics[width=\linewidth]{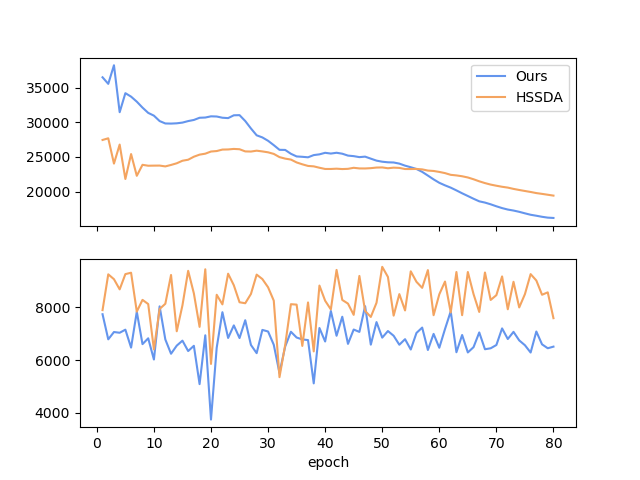}}
 \caption{The total number of incorrect pseudo-boxes on KITTI dataset. The above plot is about the number of wrong predictions of teacher model of Ours and HSSDA across training epoch. The below plot is after the pseudo-box filtering.}
 \label{fig:fp}
\end{figure}

\begin{table}[htb]
\centering
\begin{center}
\begin{tabular}{ c|cccc } 
 \hline
  \multirow{2}{*}{Model} & \multicolumn{4}{c}{1\%} \\
  & Car & Ped & Cyc & mAP \\
 \hline
 HSSDA \cite{hssda} (Reproduced) & 79.3 & 49.3 & 43.8 & 57.5 \\ 
 \hline
 + 3 channel student & 80.6 & 54.3 & 51.0 & 62.0 \\
 \hline
 + 3 channel teacher (Ours) & {\bf 82.4} & {\bf 57.0} & {\bf 56.7} & {\bf 65.4} \\
 \hline
\end{tabular}
\end{center}
\caption{Experiment of incremental channel augmentations for the student and teacher network.}
\label{tab:inc}
\end{table}

{\bf Effect of the channel augmentation.} To evaluate the effectiveness of channel augmentations separately, we conduct incremental analyses for the teacher and student models. By converting the original strong augmentation of HSSDA \cite{hssda} to strong channel augmentation for student, the total performance increased by 4.5\%. Continuously, adopting weak channel augmentation for teacher, the total performance increased about 3.4\%. Comparing with each class, car has more effect on adding channel augmentation to teacher, while minor classes (i.e., pedestrian, cyclist) take more advantage with multi-channel student.

{\bf Pseudo-box and filtering quality.} We compare the quality of pseudo-box and filtering method in Figure \ref{fig:fp}. As illustrated in the upper image, our teacher produces a noticeable number of incorrect pseudo-boxes because of the injected perturbation. However, the student TED progressively learns to extract features that remain consistent under data transformations, allowing the teacher model to predict better over time. After the filtering with two scores and channel iou consistency, false positives are significantly decreased compared to HSSDA shown at the lower plot. We keep higher quality of pseudo-boxes across training.

\section{Conclusion}
\label{sec:conc}
In this work, we demonstrate the effectiveness of introducing input channel augmentations in 3D semi-supervised object detection. We define the weak and strong channel augmentation distinguished by randomness. This strategic variation enables a tailored approach to enhance the quality of pseudo-boxes and improves model robustness and generalization. On the KITTI benchmark, we improved the state-of-the-art baseline significantly on 1\% and 2\% labeled data. 

{\bf Limitations.} While loading $N_C$ channels achieves a substantial improvement compared to SOTA works, it does come with the trade-off of increased memory demands. In addition, TED \cite{ted} requires more training time to process the $N_C$ times more data. However, note that the use of TED architecture takes a much better trade-off than naively increasing training data. More detailed analysis on this is a future work.

\vfill\pagebreak


\bibliographystyle{IEEEbib}
\bibliography{strings,refs}

\begin{thebibliography}{10}

\bibitem{temporal}
S.~Laine and T.~Aila,
\newblock ``Temporal ensembling for semi-supervised learning,''
\newblock in {\em ICLR}, 2017.

\bibitem{ts}
M.~Sajjadi, M.~Javanmardi, and T.~Tasdizen,
\newblock ``Regularization with stochastic transformations and perturbations for deep semi-supervised learning,''
\newblock in {\em NIPS}, 2016.

\bibitem{ladder}
A.~Rasmus, M.~Berglund, M.~Honkala, H.~Valpola, and T.~Raiko,
\newblock ``Semi-supervised learning with ladder networks,''
\newblock in {\em NIPS}, 2015.

\bibitem{vat}
T.~Miyato, S.-I. Maeda, M.~Koyama, and S.~Ishii,
\newblock ``Virtual adversarial training: A regularization method for supervised and semi-supervised learning,''
\newblock {\em TPAMI}, 2019.

\bibitem{pseudo}
D.-H.~Lee et~al,
\newblock ``Pseudo-label: The simple and efficient semi-supervised learning method for deep neural networks,''
\newblock in {\em Workshop on challenges in representation learning, ICML}, 2013.

\bibitem{noisy}
Q.~Xie, M.-T. Luong, E.~Hovy, and Q.V. Le,
\newblock ``Self-training with noisy student improves imagenet classification,''
\newblock in {\em CVPR}, 2020.

\bibitem{prop}
A.~Iscen, G.~Tolias, Y.~Avrithis, and O.~Chum,
\newblock ``Label propagation for deep semi-supervised learning,''
\newblock in {\em CVPR}, 2019.

\bibitem{simclrv2}
T.~Chen, S.~Kornblith, K.~Swersky, M.~Norouzi, and G.E. Hinton,
\newblock ``Big self-supervised models are strong semi-supervised learners,''
\newblock in {\em NeurIPS}, 2020.

\bibitem{mixmatch}
D.~Berthelot, N.~Carlini, I.~Goodfellow, N.~Papernot, A.~Oliver, and C.A. Raffel,
\newblock ``Mixmatch: A holistic approach to semi-supervised learning,''
\newblock in {\em NeurIPS}, 2019.

\bibitem{remixmatch}
D.~Berthelot, N.~Carlini, E.D. Cubuk, A.~Kurakin, K.~Sohn, H.~Zhang, and C.~Raffel,
\newblock ``Remixmatch: Semi-supervised learning with distribution matching and augmentation anchoring,''
\newblock in {\em ICLR}, 2020.

\bibitem{fixmatch}
K.~Sohn, D.~Berthelot, N.~Carlini, Z.~Zhang, H.~Zhang, C.A Raffel, E.D Cubuk, A.~Kurakin, and C.-L. Li,
\newblock ``Fixmatch: Simplifying semi-supervised learning with consistency and confidence,''
\newblock in {\em NeurIPS}, 2020.

\bibitem{cutout}
T.~DeVries and G.~W. Taylor,
\newblock ``Improved regularization of convolutional neural networks with cutout,''
\newblock {\em arXiv preprint arXiv:1708.04552}, 2017.

\bibitem{randaugment}
E.D. Cubuk, B.~Zoph, J.~Shlens, and Q.V. Le,
\newblock ``Randaugment: Practical automated data augmentation with a reduced search space,''
\newblock in {\em CVPR}, 2020.

\bibitem{ted}
H.~Wu, C.~Wen, W.~Li, X.~Li, R.~Yang, and C.~Wang,
\newblock ``Transformation-equivariant 3d object detection for autonomous driving,''
\newblock in {\em AAAI}, 2023.

\bibitem{hssda}
C.~Liu, C.~Gao, F.~Liu, P.~Li, D.~Meng, and X.~Gao,
\newblock ``Hierarchical supervision and shuffle data augmentation for 3d semi-supervised object detection,''
\newblock in {\em CVPR}, 2023.

\bibitem{uda}
Q.~Xie, Z.~Dai, E.~Hovy, T.~Luong, and Q.~Le,
\newblock ``Unsupervised data augmentation for consistency training,''
\newblock in {\em NIPS}, 2020.

\bibitem{stac}
K.~Sohn, Z.~Zhang, C.-L. Li, H.~Zhang, C.-Y. Lee, and T.~Pfister,
\newblock ``A simple semi-supervised learning framework for object detection,''
\newblock in {\em arXiv:2005.04757}, 2020.

\bibitem{bmvc}
H.~Choi, Z.~Chen, X.~Shi, and T-K. Kim,
\newblock ``Semi-supervised object detection with object-wise contrastive learning and regression uncertainty,''
\newblock in {\em BMVC}, 2022.

\bibitem{softteacher}
M.~Xu, Z.~Zhang, H.~Hu, J.~Wang, L.~Wang, F.~Wei, X.~Bai, and Z.~Liu,
\newblock ``End-to-end semi-supervised object detection with soft teacher,''
\newblock in {\em ICCV}, 2021.

\bibitem{sess}
N.~Zhao, T.-S. Chua, and G.H. Lee,
\newblock ``Sess: Self-ensembling semi-supervised 3d object detection,''
\newblock in {\em CVPR}, 2020.

\bibitem{3dioumatch}
H.~Wang, Y.~Cong, O.~Litany, Y.~Gao, and L.J Guibas,
\newblock ``3dioumatch: Leveraging iou prediction for semi-supervised 3d object detection,''
\newblock in {\em CVPR}, 2021.

\bibitem{detmatch}
J.~Park, C.~Xu, Y.~Zhou, M.~Tomizuka, and W.~Zhan,
\newblock ``Detmatch: Two teachers are better than one for joint 2d and 3d semi-supervised object detection,''
\newblock in {\em ECCV}, 2022.

\bibitem{proficient}
J.~Yin, J.~Fang, D.~Zhou, L.~Zhang, C.-Z. Xu, J.~Shen, and W.~Wang,
\newblock ``Semi-supervised 3d object detection with proficient teachers,''
\newblock in {\em ECCV}, 2022.

\bibitem{dds3d}
J.~Li, Z.~Liu, J.~Hou, and D.~Liang,
\newblock ``Dds3d: Dense pseudo-labels with dynamic threshold for semi-supervised 3d object detection,''
\newblock in {\em ICRA}, 2023.

\bibitem{noisedet}
Z.~Chen, Z.~Li, S.~Wang, D.~Fu, and F.~Zhao,
\newblock ``Learning from noisy data for semi-supervised 3d object detection,''
\newblock in {\em ICCV}, 2023.

\bibitem{meanteacher}
A.~Tarvainen and H.~Valpola,
\newblock ``Mean teachers are better role models: Weight-averaged consistency targets improve semi-supervised deep learning results,''
\newblock in {\em NIPS}, 2017.

\bibitem{voxelrcnn}
J.~Deng, S.~Shi, P.~Li, W.~Zhou, Y.~Zhang, and H.~Li,
\newblock ``Voxel r-cnn: Towards high performance voxel-based 3d object detection,''
\newblock in {\em AAAI}, 2021.

\bibitem{pvrcnn}
S.~Shi, C.~Guo, L.~Jiang, Z.~Wang, J.~Shi, X.~Wang, and H.~Li,
\newblock ``Pv-rcnn: Point-voxel feature set abstraction for 3d object detection,''
\newblock in {\em CVPR}, 2020.

\bibitem{kitti}
A.~Geiger, P.~Lenz, and R.~Urtasun,
\newblock ``Are we ready for autonomous driving? the kitti vision benchmark suite,''
\newblock in {\em CVPR}, 2012.

\end{thebibliography}

\end{document}